# UNL Based Bangla Natural Text Conversion – Predicate Preserving Parser Approach

**Md. Nawab Yousuf Ali, Shamim Ripon and Shaikh Muhammad Allayear**

Department of Computer Science and Engineering, East West University
Dhaka, Bangladesh

**Abstract**

*Universal Networking Language (UNL)* is a declarative formal language that is used to represent semantic data extracted from natural language texts. This paper presents a novel approach to converting Bangla natural language text into UNL using a method known as *Predicate Preserving Parser (PPP)* technique. *PPP* performs morphological, syntactic and semantic, and lexical analysis of text synchronously. This analysis produces a semantic-net like structure represented using UNL. We demonstrate how Bangla texts are analyzed following the *PPP* technique to produce UNL documents which can then be translated into any other suitable natural language facilitating the opportunity to develop a universal language translation method via UNL.

**Keywords:** *UNL, Bangla Natural Text, Predicate Preserving Parser, Morphological Analysis, Language Conversion.*

## 1. Introduction

With ever increasing number of users since its creation, Internet has evolved from monolingual into a multilingual World Wide Web of more than 1000 languages. Along with this vast number of users, the Internet usage now varies from primarily academic purposes to widespread commercial, leisure, education, entertainment etc. purposes. With this increasing pressure of providing information access without language and cultural barriers, Internet today has to deal with immense complexity of *multilinguality*. When multilingual translation is of interest, the interlingua approach allows building a system of Natural languages.

In the interlingua approach (e.g., UNL) the source language is transformed into a language independent representation, which can then be transformed into the target language. A challenge in this approach is to design a language independent intermediate representation that captures the semantic structures of all languages while remaining unambiguous. The UNL [1, 2] project is concerned about developing an intermediary language system where any written text can be converted to many languages (until now there are sixteen official languages are involved, Bangla is not yet included) through UNL and simultaneously, all text written in different languages can be converted to that particular language. Bangla is the 4th most widely spoken language with more than 250 million speakers. [3] We stress in this paper a conversion process of Bangla into UNL, so that a vast number of people can be benefitted from sharing information in the Internet.

We attempt a system to automate the generation of semantic net like expressions from text documents. The objective is to establish semantic relations between the syntactic units of a sentence to capture the syntactic and semantic attributes of words. The output of the system is a set of UNL Expressions, which are binary relations among disambiguated words along with speech acts attributes attached to these disambiguated words.

In the rest of the paper, we first briefly outline the semantic based approach to processing document in UNL along with UNL's translation scheme. In the following section, we introduce the analyzer system, Enconverter, which is used to convert Bangla sentences into UNL form (Section 2). We then describe the *PPP* technique and explain how it is being applied in our experiments in Section 4. Section 4 describes how a Bangla sentence is encoded. The analysis methods of Bangla sentences in favour of UNL are demonstrated in Section 6. Finally, we conclude our paper summarizing our work and outlining our future plan in Section 7.

## 2. UNL

The UNL [1] has been defined as a digital meta language for describing, summarizing, refining, storing and disseminating information in a machine independent and human language neutral form. It represents information, i.e. meaning, sentence by sentence. Each sentence is represented as a hypergraph, where nodes represent concepts and arcs represent relation between concepts. This hypergraph is also represented as a set of directed binary relations between the pair of concepts present in the sentence. Concepts are represented as character-strings

called Universal Words (UWs). Knowledge within a UNL document is expressed in three dimensions:
i) *Universal Words (UWs)*: Word knowledge is expressed by Universal Words which are language independent. UWs constitute the UNL vocabulary and the syntactic and semantic units that are combined according to the UNL laws to form UNL expressions. They are tagged using restrictions describing the sense of the word in the current context. For example, *drink(icl>liquor)* denotes the noun sense of drink restricting the sense to a type of liquor. Here *icl* stands for inclusion and form an *is-a* relation as in semantic nets [4].
ii) *Relation Labels*: Conceptual knowledge is captured by the relationship between Universal Words (UWs) through a set of UNL relations. For example, *Human affects the environment* is described in the UNL expression as,
*{unl}*
*agt(affect(icl>do).@present.@entry:01,*
   *human(icl>animal).@pl)*
*obj(affect(icl>do).@present.@entry:01,*
   *environment(icl>abstract thing).@pl)*
*{/unl}*
where, *agt* means the agent and *obj* means object. The terms *affect(icl>do)*, *human(icl>animal)* and *environment(icl>abstract thing)* are the UWs denoting concepts.
iii) *Attribute Labels*: Speaker's view, aspect, time of event, etc. are captured by UNL attributes. For instance, in the above example, the attribute *@entry* denotes the main predicate of the sentence, *@present* denotes the present tense, *@pl* is for the plural number and *:01* represents the scope ID.

A UNL expression can also be represented as a graph. For example, the UNL expressions and the UNL graph for the sentence, *I went to Malaysia from Bangladesh by aeroplane to attend a conference,* are shown in Fig. 1.

In the Fig.1, *agt* denotes the agent relation, *obj* the object relation, *plt* the place relation denoting the place to go, *plf* is also a place relation that denotes the place from, *pur* states the purpose relation, whereas *met* is for method relation.

UNL expressions provide the *meaning content* of the text. Hence, search could be carried out on the meaning rather than on the text. This of course means developing a novel kind of search engine technology. The merit of such a system is that the information in one language can be stored in multiple languages.

*{unl}*
  *agt(go(icl>move>do,plt>place,plf>place,*
    *agt>thing).@entry.@past,i(icl>person))*
  *plt(go(icl>move>do,plt>place,plf>place,*
    *agt>thing).@entry.@past,*
    *malaysia(iof>asian_country>thing))*
  *plf(go(icl>move>do,plt>place,plf>place,*
    *agt>thing).@entry.@past,*
    *bangladesh(iof>asian_country>thing))*
  *met(go(icl>move>do,plt>place,plf>place,*
    *agt>thing).@entry.@past,*
    *aeroplane(icl>heavier-than-air_craft>thing,*
    *equ>airplane))*
  *obj:01(attend(icl>go_to>do,agt>person,*
    *obj>place).@entry,*
    *conference(icl>meeting>thing).@indef)*
  *pur(go(icl>move>do,plt>place,plf>place,*
    *agt>thing).@entry.@past,:01)*
*{/unl}*

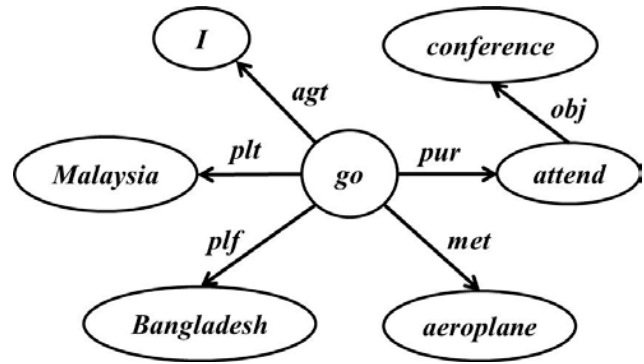

Fig. 1  UNL expression and UNL graph.

## 3. The Analyzer System: Enconverter

To convert Bangla sentences into UNL form, we use EnConverter (*EnCo*) [5], a universal converter system provided by the UNL project [6]. The *EnCo* is a language independent parser, which provides a framework for morphological, syntactic and semantic analysis synchronously. Natural Language texts are analyzed sentence by sentence by using a knowledge rich lexicon and by interpreting the analysis rules. It scans an input sentence from left to right. When an input string is scanned, all matched morphemes with the same starting characters are retrieved from the dictionary and become the candidate morphemes according to the priority rule in order to build a syntactic tree and the semantic network for the sentence.

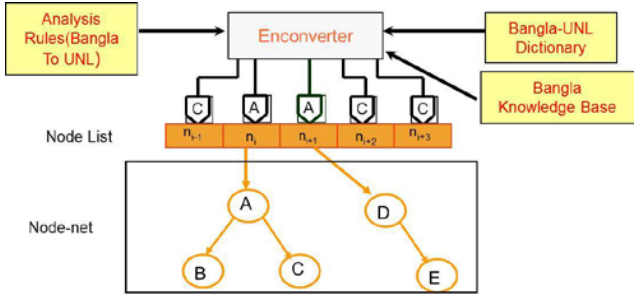

Fig. 2  Structure of Enconverter.

The left character string is scanned from the beginning according to the applied rule. Fig. 2 shows a structured mult-headed Turing Machine. It moves back and forth over the *Node List*, which contains words of the input sentence. In the figure, "A" indicates an *Analysis Window*, "C" indicates a *Conditional Window*, and $n_n$ indicates an *Analysis Node*. The *EnCo* uses the Conditional Windows for checking the neighboring nodes on both sides of the Analysis Windows} in order to check whether the neighboring nodes satisfy the conditions to apply an *Analysis Rule*. The Analysis Windows are used to check two adjacent nodes in order to apply one of the Analysis Rules. If there is an applicable rule, *EnCo* adds Lexical attribute to or deletes Lexical attributes from these nodes, and creates a partial syntactic tree and UNL network according to the type of rule.

3.1 Lexicon

The lexicon consists of mapping of Bangla words to Universal Words and lexical-semantic attributes describing the words. The format [1] of an entry in the dictionary is as follows,

[HW] {ID}"UW"(ATTRIBUTE1, ATTRIBUTE2,…)<FLG, FRE, PRI>

where,

|  |  |
|---|---|
| HW: | Head Word (Bangla Word) |
| ID: | Identification of Head Word |
| UW: | Universal Word |
| ATTRIBUTE: | Attribute of the HW |
| FLG: | Language Flag (e.g. B for Bangla) |
| FRE: | Frequency of Head Word |
| PRI: | Priority of Head Word |

Some examples of Dictionary entries for Bangla language texts are presented in Fig. 3.

```
[মাছ]{}"fish (icl>animal>animate thing)"
          (N, ANI,SG, CONCRETE) <B,0,0>
[সে]{}"he(icl>person)" (PRON,MALE,ANI, 3SG,
                              HPRON) <B,0,0>
[সুন্দর]{}"beautiful(icl>state)"(ADJ)<B,0,0>
```

Fig. 3  Example entries of Bangla Word Dictionary.

In the dictionary entries shown in Fig. 3, *N* stands for Noun, *ANI* stands for animate object, *SG* stands for singular, *CONCRETE* for concrete object, *PRON* for pronoun, *MALE* for gender is male, *3SG* for third person singular, *HPRON* for human pronoun, *ADJ* for adjective and the entry *B* stands for Bangla.

The attributes in the lexicon are collectively called *Lexical Attribute*. The syntactic attributes include the word category, such as, `noun', `verb', `adjectives', etc. and attributes like `person' and `number' for nouns, and `tense' information for verbs.

## 4. Predicate Preserving Parser

In predicate preservation, the focus is to attempt to locate the main predicate of the sentence at every step of the analysis. At every combination or modification operation, a node is deleted from the Node-list, if there is a suitable candidate node for the predicate of the sentence or the clause is under the other Analysis Window. For example, if there is a proper verb (neither an auxiliary nor a modal) under the Left Analysis Window (LAW), and there is a noun, which is an object, under Right Analysis Window (RAW), the noun is then deleted preserving the verb. Similarly, if there is an auxiliary verb like *is* is under the LAW and there is a noun under the RAW, then the noun is preserved.

Both Morphological Analysis and Decision Making are taken place at every step while the *EnCo* scans the input sentence from left to right. Morphology is concerned when words are formed from basic sequences of morphemes. It acts as the crossroads between phonology, lexicon, syntax, semantics and context. There are two types of morphology: (i) *Inflectional Morphology* and (ii) *Derivational Morphology*}. Inflectional morphology [7, 8] is required to identify the actual meaning of the word.

Every word is derived from a root word. A root word may have various transformations. It happens because of adding different morphemes with it as suffixes. As a result, the meaning of the word varies for various transformations.

For example, if we consider a Bangla word 'খা' (eat) as a root word then after adding 'বে' {be}' we get a word খা-বে (will eat). Similarly, 'খা-ইতেছে' (is eating), 'খে-য়েছিল'(ate) etc. Our approach for implementing Inflectional Morphology is different from the conventional approach. The Universal Words defined in the lexicon have word roots specified as Universal Words followed by appropriate restrictions while the headword has the Longest Common Lexical Unit (LCLU) of all the possible transformations of the word. For example, the dictionary entry for the verb খা (eat) is:

[খা]{}"*eat(icl>consume>do,agt>living_thing, obj>concrete_thing)*"(ROOT,VEND,VEG1,#AGT,#OBJ)<B,0,2>

where, খা (eat) is the LCLU for খাবে (will eat), খাইতেছে (is eating), খেয়েছিল (ate) etc. The dictionary entries for these morphemes are as follows,

[বে]] {}"" (*VI, 2P, PRGR, SD*)
[ইতেছে] {} "" (*VI, 2P, FUT, SD*)

where, *VI* stands for verbal inflexion, *2P* is for second person, *PRGR* is for progress which means continuous tense, *SD* is for standard language text and *FUT* denotes future tense. *EnCo* selects the longest matched entry from the lexicon, starting from the first character. When EnCo consults the dictionary for a particular morpheme, e.g., ইতেছে (eating), it will be able to retrieve the LCLU (খা here) that is available in the dictionary. The analysis rules then take control, and look ahead and signal that there is a verb suffix ahead and complete the morphological analysis.

In the case of decision making according to the lexical attributes of the nodes under the two Analysis Windows, the parser decides whether (i) the nodes are to be combined into a single headword, or (ii) a relation is to be set up between them, or (iii) an UNL attribute is to be generated. While combining or modifying the two nodes, one of the nodes is deleted from the node-list. When multiple rules can be applied, we need a mechanism to select a rule. Priorities can be assigned to the rules similar to expert system. We adopted the following prioritizing strategy,

1) Morphological analysis rules have the highest priority. Obviously, unless we have the morphed word we cannot decide the part of speech of the word and its relation with the adjacent words.
2) Rules for dealing with specific constructs are given higher priority than those for general sentence structures. For instance, rules for clausal and passive sentences are given higher priority, so that while analyzing clausal or passive sentences a general rule-eligible to be applied-does not fire.
3) Right shift rules which facilitate right movement when there is nothing else to do are given the lowest priority. For example, when the LAW is on SHEAD (sentence start marker) and the RAW (Right Analysis Window) is on the subject (N), no rule other than the right shift is applicable.
4) Composition rules are usually given less priority than modification rules. The former ultimately resolve relations while the latter change the properties of the nodes under the AWs.

## 5. Bangla Sentence Encoding

The encoding process will be performed by shift/reduce parsing [9]. Both Bangla and Hindi (Indian Language) are derived from Sanskrit language. Hence, Bangla and Hindi languages have a lot of syntactic similarities. The conversion of UNL from Hindi language has been shown in [10]. Due to the syntactic similarities, this paper follows a similar approach as for Hindi language. In order to explain the encoding steps we give an example of the analysis of a simple assertive sentence. Assertive simple sentences have only one main clause. We assume that analysis rules and the dictionary of Bangla to UNL are given to the analyser system *EnCo*. Consider the following simple English sentence *"It is very hot in Dhaka today"*. The corresponding Bangla text and its word by word meaning in English is given as,

ঢাকায়    আজ খুব গরম
<u>Dhaka-in</u> <u>today</u> <u>very</u> <u>hot</u>

The node list is enclosed within "<<" and ">>". The analysis window is enclosed within "[" and "]". The nodes delimited by "/" are those explored and fixed by the system. Morphological analysis is carried out is a series of steps.

1. /<</ [ঢাকা] / [য়] / "আজ খুব গরম" />>/
2. /<</[ঢাকা য়]/ "আজ খুব গরম" />>/
3. /<</ ঢাকা য় / আজ / [খুব] /[গরম] />>/
4. /<</ ঢাকা য় / [আজ] / [গরম] />>/
5. /<</[ঢাকা য়] / [গরম] />>/
6. /<</ [গরম] />>/

**Step 1:** The noun ঢাকা (*Dhaka*) of the type *PLACE* and case maker য় (in) are combined and an attribute *PLC*

is added to the noun to indicate that *plc* relation can be made between the main predicate of the sentence and this noun.

- **Step 2:** In this step, the system right shifts here because there is no combination or modification rule between noun and adverb.
- **Step 3:** The system recognizes গরম (*hot*) as a predicate of the sentence. So, it generates *man* relation between the adverb খুব (*very*) and the predicate গরম (*hot*). The adverb খুব (*very*) is deleted afterwards.
- **Step 4:** Then the analyzer looks ahead further right beyond the noun phrase আজ (*today*) to get the predicate গরম (*hot*). A *tim* relation is created between আজ (*today*), and গরম (*hot*) and finally, আজ (today) is deleted.
- **Step 5:** By using *PLACE* attribute of the noun ঢাকা (*Dhaka*), the system generates *plc* relation between the noun ঢাকা and the predicate গরম.
- **Step 6:** A right shift at this point brings the Sentence Tail (STAIL) under the LAW and thus signals the end of analysis. This right shift rules also attaches the attribute *@entry* to the last word left in the Node-list and thus the predicate গরম (*hot*) is preserved till the end.

The UNL output of the corresponding sentence is as follows:

[S]
*man (hot (icl>state).@entry.@present, very (intensifier))*
*tim(hot(icl>state).@entry.@present, today(icl>period))*
*plc (hot (icl>state).@entry.@present, Dhaka (icl>place))*
[/S]

Other types of sentences, e.g., interrogative, exclamatory, can also be encoded in a similar fashion.

## 6. Analysis of Bangla Sentences

The rule base of the Bangla Analyzer (BA) is broadly divided into three categories, namely, *morphological rules*, *composition rules*, and *relation resolving rules*. Bangla language has a rich morphological structure. Information regarding person, number, tense and gender can be extracted from the morphology of nouns, adjectives and verbs. A considerable number research work has been done on morphology [11-13].

Here we describe various Bangla language phenomena handled by the system. Bangla is a null subject language. This means that it allows the syntactic subject to be absent. For example, consider the following valid Bangla sentence along with its word by word and English translation.

Bangla: কাজ করছি
<u>Working</u> <u>am</u>
English: am working

The system makes the implicit subject explicit in the UNL expressions. The following UNL expression is produced by the system,

[S]
*agt(work(icl>do).@entry.@present. @progress, I(icl>person))*
[/S]

The following example shows how the system can handle a limited number of anaphora. Consider the following Bangla sentence along with its English translation,

Bangla: রোজী তার বইটি শহিদা-কে দিয়ে-ছে
<u>Rozi</u> <u>her</u> <u>book</u> <u>Shahida-to</u> <u>given-has</u>
English: Rozi has given her book to Shahida

The corresponding UNL expression of the sentence generated by the system is,

[S]
*pos(book(icl>publication),Roji(icl>person):01)*
*ben(give(icl>do.@entry.@present.*
            *@pred,Shohida(icl>person))*
*obj(give(icl>do.@entry.@present.*
            *@pred,book(icl>publication))*
*agt(give(icl>do.@entry.@present.*
            *@pred,Roji (icl>person):01)*
[/S]

The resolution of the anaphora is apparent from the fact that the UW *she(icl>person)* for *her* is replaced by *Roji(icl>person)* in the *pos* relation.

Bangla is a relatively free word-ordered language; the same sentence can be written in more than one way by changing the order of the words. The following three sentences have the same meaning but the word order of each sentence is different.

Bangla:
a) সে কোথায় যা-ইতেছে?
   He where going is
b) কোথায় সে যা-ইতেছে?
   Where he going is
c) কোথায় যা-ইতেছে সে?
   Where going is he
English: Where is he going?

The UNL output for all cases is as follows,

```
[S]
plc(go>do).@entry.@interrogative.@present.
        @pred.@progress,where(icl>place):01)
agt(go>do).@entry.@interrogative.@present.
        @pred.@progress,se(icl>male):02)
[/S]
```

For the sentence in (a), first the rule for generating *plc* relation between '*where*' and '*going is*' is fired, followed by the rule for generating *agt* relation between '*He*' and '*going is*'. In sentence (b), first *agt* then *plc* are resolved. In sentence (c), first a rule is used to exchange the positions of '*going is*' and '*he*'. After that the relations are set up by firing the required rules.

One of the major difference between Bangla and English is that a single pronoun e.g. সে in Bangla can be represented in English '*He*' or '*She*'. So, in UNL-Bangla dictionary this single word is included in two different dictionary entries as follows,

[সে] "He(icl>person)" (PRON, MALE, 3SG, ANI, HPRON)
[সে] "She(icl>person)" (PRON, FEMALE, 3SG, ANI, HPRON)"

## 7. Conclusions

This paper presented a unique technique named *Predicate Preserving Parsing* to translate a natural language text, in our case it is Bangla, into UNL expressions. The UNL expressions preserve the semantic structure of the natural language texts and can be converted into any other language with the help of language specific analysis and generation rules, and dictionary entries. The encoding and decoding of a language to and from UNL are carried out by EnConverter and DeConverter tools provided by UNL foundation. We implemented our technique by keeping the main predicate of the analyzing sentence at each step of the conversions. Our technique is working well for conversion of any simple assertive sentence into UNL expressions. We have done several experiments of numerous combinations of nouns/pronouns/verbs over simple sentences.

The conversion of complex and compound sentences are yet to be completed. Currently, we are investigating on the analysis and generation rules of these types of sentences. It is essential to check the effectiveness of the interlingua approach that has been used here, i.e., determining whether the meaning of the source language is being conveyed properly, so that a native speaker of the target language accepts it as natural. A similar work has been carried out in [14] for Hindi language. The approach could be followed for Bangla as well. A long term plan is to investigate the use of UNL as a knowledge representation scheme and use this knowledge for various purposes, like text mining, document classification and other knowledge intensive tasks.

**Md. Nawab Yousuf Ali** obtained PhD in Computer Science and Engineering from the Department of Computer Science and Engineering, Jahangirnagar University, Dhaka, Bangladesh. He is serving as an Asst. Prof. at East West University, Bangladesh. He is interested in Natural Language Processing, UNL, Bangla text conversion to UNL. He has published several research papers in national and International journal and conferences.

**Shamim Ripon** received his PhD in computer Science from University of Southampton, UK. He is an Asst. Prof. at East West University, Bangladesh where he also leads Software Engineering and Formal Method Research group. His primary research interests are focused on Formal Methods, Requirement Engineering, Software Product Line and Natural Language Processing.

**Shaikh Muhammad Allayear** completed his MSc and PhD in Computer Science & Engineering from Anyang University, South Korea. He was also a part time lecturer of Department of Computer Science & Engineering, Anyang University. Currently he is an Assistant Professor of the Department of Computer Science and Engineering (CSE), East West University, Dhaka, Bangladesh. His research interest includes Network Storage, Internet Computing, Parallel & Distributed System, High Performance Computing, Mobile Computing, and Information Security.